\documentclass[12pt,halfline,a4paper]{ouparticle}
\usepackage[utf8]{inputenc}
\usepackage{multirow}
\usepackage{tabularx}
\usepackage{longtable} 
\usepackage{rotating}
\usepackage{colortbl}
\usepackage{natbib}
\usepackage{graphicx}
\usepackage{hyperref}
\usepackage{todonotes}

\begin{document}

\title{Unlocking the potential of deep learning for marine ecology: overview, applications, and outlook\footnote{All authors have an equal contribution. Authors are ordered alphabetically.}}

\author{%
\name{Morten Goodwin}
\address{Centre for Artificial Intelligence Research,\\ University of Agder, 4604 Kristiansand, Norway}
\email{morten.goodwin@uia.no}
\and
\name{Kim Tallaksen Halvorsen}
\address{Institute of Marine Research, Ecosystem Acoustics Group\\
Nye Flødevigveien 20, 4817 His, Norway}
\email{kim.halvorsen@hi.no}
\and
\name{Lei Jiao}
\address{Centre for Artificial Intelligence Research,\\ University of Agder, 4604 Kristiansand, Norway}
\email{lei.jiao@uia.no}
\and
\name{Kristian Muri Knausgård}
\address{Top Research Centre Mechatronics,\\University of Agder, 4879~Grimstad, Norway}
\email{kristianmk@ieee.org}
\and
\name{Angela Helen Martin}
\address{Center for Coastal Research, University of Agder,\\
4604, Kristiansand, Norway}
\email{angelahmartin@protonmail.com}
\and
\name{Marta Moyano}
\address{Center for Coastal Research, University of Agder,\\
4604, Kristiansand, Norway}
\email{marta.moyano@uia.no}
\and
\name{Rebekah A. Oomen}
\address{Center for Coastal Research, University of Agder,\\
4604, Kristiansand, Norway}
\address{Centre for Artificial Intelligence Research, University of Agder,\\ 
4604 Kristiansand, Norway}
\address{Center for Ecological and Evolutionary Synthesis, University of Oslo,\\
0371, Oslo, Norway\\
+47 47 71 44 17}
\email{rebekahoomen@gmail.com}
\and
\name{Jeppe Have Rasmussen}
\address{Center for Coastal Research, University of Agder,\\
4604, Kristiansand, Norway}
\address{Centre for Artificial Intelligence Research, University of Agder,\\ 
4604 Kristiansand, Norway}
\email{jeppe.h.rasmussen@uia.no}
\and
\name{Tonje Knutsen Sørdalen}
\address{Center for Coastal Research, University of Agder,\\
4604, Kristiansand, Norway}
\email{tonjesordalen@gmail.com}
\and
\name{Susanna Huneide Thorbjørnsen}
\address{Institute of Marine Research, Flødevigen,\\
Nye Flødevigveien 20, 4817 His, Norway}
\address{Center for Coastal Research, University of Agder,\\
4604, Kristiansand, Norway}
\email{susanna.thorbjornsen@gmail.com}}

%
%









\abstract{
The deep learning revolution is touching all scientific disciplines and corners of our lives as a means of harnessing the power of big data. Marine ecology is no exception. These new methods provide analysis of data from sensors, cameras, and acoustic recorders, even in real time, in ways that are reproducible and rapid. Off-the-shelf algorithms can find, count, and classify species from digital images or video and detect cryptic patterns in noisy data. Using these opportunities requires collaboration across ecological and data science disciplines, which can be challenging to initiate. To facilitate these collaborations and promote the use of deep learning towards ecosystem-based management of the sea, this paper aims to bridge the gap between marine ecologists and computer scientists. We provide insight into popular deep learning approaches for ecological data analysis in plain language, focusing on the techniques of supervised learning with deep neural networks, and illustrate challenges and opportunities through established and emerging applications of deep learning to marine ecology. We use established and future-looking case studies on plankton, fishes, marine mammals, pollution, and nutrient cycling that involve object detection, classification, tracking, and segmentation of visualized data. We conclude with a broad outlook of the field’s opportunities and challenges, including potential technological advances and issues with managing complex data sets.}

\date{\today}


\maketitle

\section{Introduction}
Marine ecosystems are complex, highly diverse, and productive, providing renewable resources to a growing human population. At the same time, the oceans are particularly sensitive to and impacted by anthropogenic stressors~\Citep{Antao2020}. As such, the scientific community strives to deliver up-to-date information about the state of marine ecosystems so that management decisions are well-informed. Ideally, such decisions use ecosystem-based management~(EBM) approaches to preserve ecosystem health and productivity while allowing appropriate human use. EBM is especially relevant in densely populated coastal areas. During this period of rapid environmental change, EBM requires researchers to track ecological change and critical events when, and not well after, they occur. Fortunately, technological developments in observation methods have provided ecologists with a range of new tools for obtaining vast amounts of data from marine ecosystems over the last couple of decades. These include high-end cameras, echo sounders, and hydrophones, combined with various sensors to measure environmental parameters. Researchers can attach such technologies to cabled observatories or static rigs to assess temporal dynamics, or remotely or autonomously operated vehicles to evaluate spatial variability. However, because these technologies can produce an unprecedented amount of data, which has traditionally required manual processing, ecologists may be reluctant to adopt them as an alternative or supplement to traditional sampling techniques. For example, using traditional gear~(e.g., nets and traps) to assess the abundance of fish has been an established sampling technique for centuries and is still used today. These methods are efficient for manual data handling and straightforward: as soon as the fish are caught, counted, and the data punched, it can be analyzed by the researchers.
On the other hand, detecting and counting fish with cameras is less destructive to animals and habitat, provides a temporal dimension to the collected data, allows researchers to observe behaviour of animals and habitat use, and often provides a more representative estimate of species diversity and relative abundance~\Citep{Bacheler2017}. However, extracting all of this information from videos manually is a laborious task. Thus, automating this step would undoubtedly encourage more fish biologists to use cameras for data collection. 

Many diverse fields of research are undergoing rapid change due to advances in the use of artificial intelligence~(AI) for data interpretation. AI offers fast and accurate analysis of the large volumes of data collected by sensors, cameras, and other observation technologies. Off-the-shelf algorithms can now, with high precision, find, count, and classify organisms from digital images and real-time video,~\Citep{knausgaard2021temperate,lopez2020sensors} and detect cryptic patterns in noisy images or acoustic data~\Citep{weinstein2018}. An increasing number of marine ecologists embrace this opportunity, yet initiating collaborations across ecological and data science disciplines can be challenging for several reasons. First, transferring the necessary information to start a project between an ecologist and a computer scientist can be a steep learning curve because knowledge barriers and field-specific jargon can cloud otherwise fruitful discussions and halt progression. Secondly, ecologists unfamiliar with AI may not be aware of the opportunities available to address a particular problem. Before an ecologist approaches an AI expert, they may need to know about the possibilities and limitations of AI for the task at hand, how to prepare and annotate data sets, and what information to provide the computer scientist to enable identification of the best AI method for the task at hand. Meanwhile, before advising on the possibilities, the computer scientist may find it challenging to understand the underlying ecological question, the data and its inherent variability/noisiness, how it is categorized, and what level of accuracy is needed. Thus, substantial investment in the interdisciplinary partnership is required in order to achieve a common understanding.
	
This paper aims to bridge the gap between marine ecologists and computer scientists to expedite the initial stages of collaboration. Here, we provide insight into the most popular and suitable AI techniques for ecological data analysis and describe technical concepts in plain language. AI is a general term referring to any artificial intelligence technique that can solve a complicated problem~\Citep{goodwin2020ai,russell2002artificial}. We focus on applicable and well-used methods, namely deep neural networks~(DNNs), synonymous with ``deep learning'', and learning with supervision~(supervised machine learning). Supervised learning requires algorithms to be presented with datasets that have been labeled with accurate information on the region of interest, for example the presence or location of known species, objects, or sound. The algorithms learn to associate the labels with the examples~\Citep{christin2019}. With enough training material, the algorithms can produce models that automatically recognize and identify new and unseen examples in other datasets without the need for new labels~\Citep{lecun2015deep}. One of the biggest challenges for supervised learning is the demand for a large, labeled training dataset of sufficient quality to achieve high accuracy~\Citep{malde2020machine,beyan2020setting}. Close collaboration between ecologists and computer scientists would likely facilitate and accelerate the dedicated effort required to collect and label representative datasets~\Citep{weinstein2018, Schneider_2019, beyan2020setting}. 

This paper is organized as follows: Section~\ref{AIoverview} summarizes popular deep learning tools relevant for ecologists and explains standard AI terms. Section~\ref{sec:Established} describes three cases where AI has been applied to ecological data, namely, fish detection, classification, and tracking in underwater videos; image-based analysis for plankton monitoring; and acoustic monitoring of whales. Deep learning ecology research is not limited to these cases and we are confident that the deep learning toolset will have an even greater impact on emerging research areas in marine ecology. Therefore, section \ref{sec:roadahead} continues with four case studies where we see potential for deep learning to make an essential impact, including individual re-identification of fish using unique patterns; analysing fish vocal communication to understand mating behavior; ghost fishing gear detection; and determining the ecological functions of fishes. Finally, in section \ref{sec:Discussion}, we discuss technological advances, complexity in data, and acceleration of data collection and labelling through open-source approaches.

\begin{sidewaystable*}
\caption{Machine learning approaches to ecological data applied~(green) or explored~(blue) in the case studies~(C1-C7), and some alternatives~(orange). Grey cells indicate no added benefit to using that approach for the task. Approaches: (Ap A): One label per region of interest, (Ap B): One label per image, (Ap C): Pixel-wise segmentation, (Ap D): Ground truth spectrograms with labeled region of interest, (Ap E): Labelled spectrograms with regions of interest, and (Ap F): Segmented time series data.
}\label{tab:CaseTable}
%
\definecolor{ColorEstablished}{HTML}{C5E0B4}
\definecolor{ColorEmerging}{HTML}{FFCA64}
\definecolor{ColorAlternatives}{HTML}{81CFF5}
\definecolor{ColorNoBenefit}{HTML}{B7B7B7}
\renewcommand{\arraystretch}{1.25}
\small
\begin{center}
\begin{tabular}{ | m{9em} | m{8em} | m{8em} | m{8em} | m{8em} | m{8em} | m{8em} | } 
\hline
\multirow{2}{9em}{Case studies}& \multicolumn{2}{c|}{Object detection} & \multicolumn{2}{c|}{Classification} & \multicolumn{2}{c|}{Segmentation} \\ 
\cline{2-7}
& When to use & Possible method & When to use & Possible method & When to use & Possible method \\ 
\hline
Fish and species counting (C1) & \cellcolor{ColorEstablished!100} Images with 1+ fish & \cellcolor{ColorEstablished!100} YOLO with \mbox{Ap A} & \cellcolor{ColorAlternatives!100} Images with 0/1 fish / species & \cellcolor{ColorAlternatives!100} Squeeze-and-excitation with \mbox{Ap B} & \cellcolor{ColorNoBenefit!100} Outlines of 1+ regions wanted & \cellcolor{ColorNoBenefit!100} R-CNN with \mbox{Ap C}\\
\hline
Plankton analysis (C2) & \cellcolor{ColorEstablished!100} Images with 1+ organisms & \cellcolor{ColorEstablished!100} YOLO with \mbox{Ap A} & \cellcolor{ColorEstablished!100} Images with single organisms & \cellcolor{ColorEstablished!100} CNN with \mbox{Ap B} & \cellcolor{ColorEstablished!100} Images with single organism (morphology) & \cellcolor{ColorEstablished!100} R-CNN with \mbox{Ap C}\\
\hline
Marine bioacoustics (C3) & \cellcolor{ColorEstablished!100} Spectrograms with 1+ calls & \cellcolor{ColorEstablished!100} R-CNN with \mbox{Ap D} & \cellcolor{ColorEstablished!100} Spectrograms with 1/0 calls & \cellcolor{ColorEstablished!100} CNN with \mbox{Ap F} & \cellcolor{ColorAlternatives!100} Separation for 0+ calls in time series & \cellcolor{ColorAlternatives!100} RNN with \mbox{Ap G} or transformer with \mbox{Ap G}\\
\hline
Re-identification in fish populations (C4) & \cellcolor{ColorAlternatives!100} Images with 1+ fish & \cellcolor{ColorAlternatives!100} YOLO with \mbox{Ap A} & \cellcolor{ColorEstablished!100} Images with 0/1 individuals & \cellcolor{ColorEstablished!100} CNN with \mbox{Ap B} & \cellcolor{ColorAlternatives!100} Images with fish outlined & \cellcolor{ColorAlternatives!100} R-CNN with \mbox{Ap C}\\
\hline
Fish vocal communication (C5) & \cellcolor{ColorEstablished!100} Spectrograms with 1+ individual calls & \cellcolor{ColorEstablished!100} R-CNN with \mbox{Ap D} & \cellcolor{ColorEstablished!100} Spectrograms with 1/0 calls & \cellcolor{ColorEstablished!100} CNN with \mbox{Ap F} & \cellcolor{ColorNoBenefit!100} Separation for 0+ calls in time series & \cellcolor{ColorNoBenefit!100} RNN with \mbox{Ap G} or transformer with \mbox{Ap G}\\
\hline
Ghost fishing gear detection (C6) & \cellcolor{ColorEmerging!100} Images with 1+ gear & \cellcolor{ColorEmerging!100} R-CNN with \mbox{Ap A} & \cellcolor{ColorAlternatives!100} Images with 0/1 gear & \cellcolor{ColorAlternatives!100} CNN with \mbox{Ap B} & \cellcolor{ColorNoBenefit!100} Areas with partially dissolved fishing nets& \cellcolor{ColorNoBenefit!100} R-CNN with Ap C\\
\hline
Carbon cycling by fish (C7) & \cellcolor{ColorEmerging!100} Images with 1+ life processes & \cellcolor{ColorEmerging!100} R-CNN with \mbox{Ap A} & \cellcolor{ColorEmerging!100} Images with 0/1 life processes & \cellcolor{ColorEmerging!100} CNN with \mbox{Ap B} & \cellcolor{ColorEmerging!100} Images or video with moving processes & \cellcolor{ColorEmerging!100} YOLO with \mbox{Ap C}\\
\hline
\end{tabular}
\end{center}
\end{sidewaystable*}

\section{A non-comprehensive review of deep learning}\label{AIoverview}

\begin{figure}[h!t]

\includegraphics[width=1\textwidth]{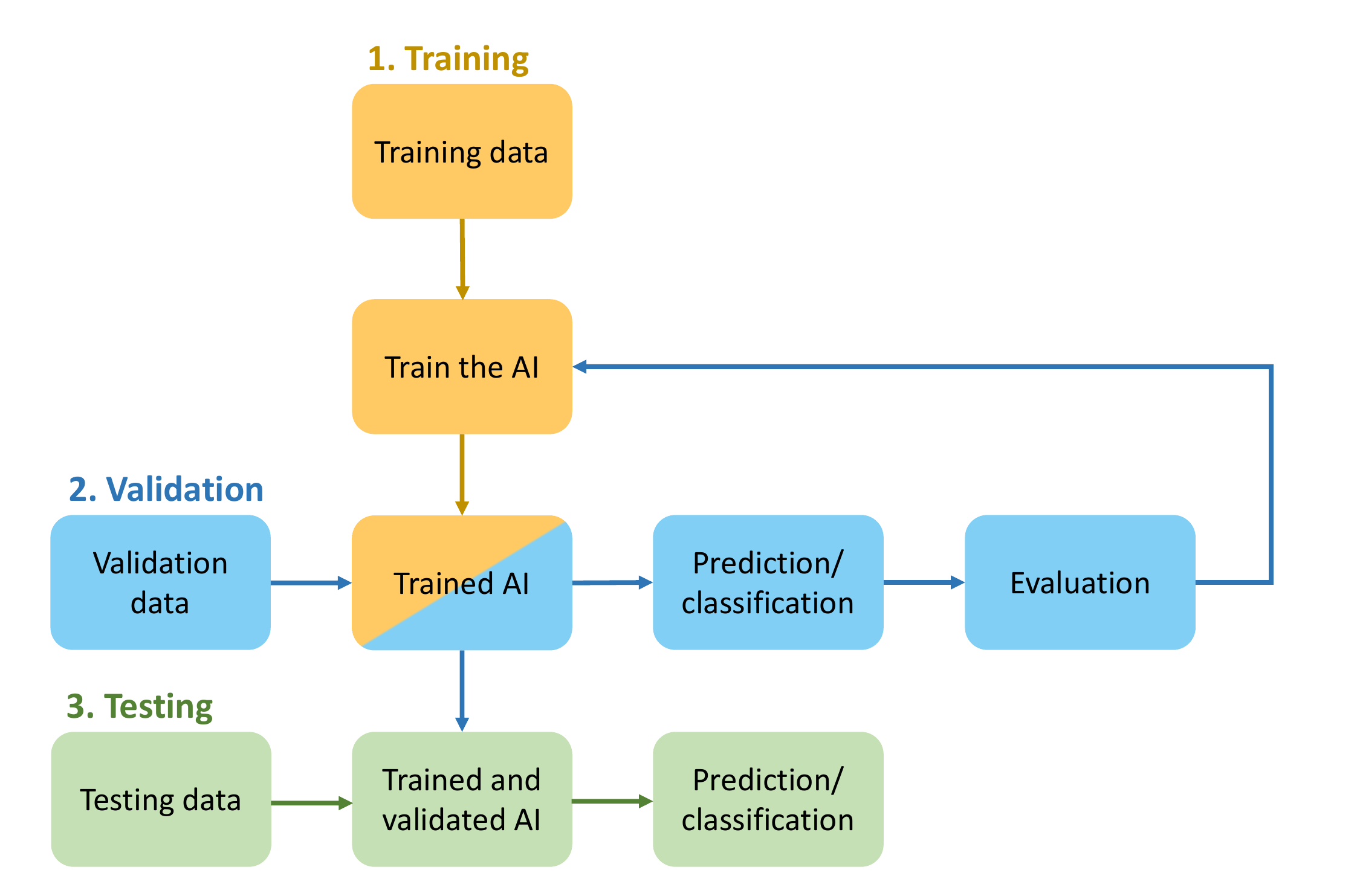}
\caption{The workflow of AI based strategies. (1) The (yellow) column illustrates the training phase, in which labeled data is used to train the AI algorithm. (2) The first row (blue) shows that the performance of the trained AI is evaluated using a validation data set and the AI algorithm may be updated and refined in this process. (3) The bottom row (green) shows the application phase, using the AI on a test data set once the training and testing are completed.}\label{fig:AI}
\end{figure}

AI is a broad concept, but the most commonly applied technique is machine learning. Machine learning is a set of algorithms that learn from an environment containing data such as images. The most common AI approach used in biology is supervised learning, which is when the data are labeled or categorized so that the algorithms can learn from the data. Conversely, unsupervised learning is when algorithms do not use labelled data but, instead, learn data structures that are reinforced when the algorithms continuously interact with an environment, such as playing a board game.  Figure \ref{fig:AI} illustrates the overall procedure for training and application of AI with supervised learning.    

Among the most popular and widely used AI algorithms are the family of artificial neural networks. A neural network is a set of human brain-inspired networks with artificial neurons and synapses that are trained to approximate an external function, typically mapping from input data~(e.g., images) to labeled values or categories~(e.g., classes). A neural network consists of a layer of input neurons connected to the input data and a layer of output neurons mapping to the values or categories to be predicted. It is common to have layers between the input and output, which are referred to as hidden layers. When a network has more than one hidden layer, it is referred to as deep learning~(DL) or a deep neural network (DNN). 

Neural networks, especially DL, are the go-to machine learning approach for categorizing and recognizing images and sound data. These techniques have won numerous pattern recognition and machine learning competitions for image and sound analytics~\Citep{tessler2017deep,schmidhuber2015deep}. In recent years, DL has become the predominant analytical technology in many domains, including health~\Citep{esteva2019guide}, customer evaluation~\Citep{lessmann2019targeting}, and crisis management~\Citep{lazreg2019iterative,ben2019not}. Aquatic ecology has experienced the early stages of the same shift, where object detection and semantic segmentation are being used to identify and locate marine species in raw images, videos, and audio recordings for the purpose of species~\Citep{knausgaard2021temperate} and individual~\Citep{bogucki2019applying} classification, and to quantify abundance. Despite the domination of deeper over more shallow neural networks, there is no need to employ DL models exclusively. Depending on the complexity and the nature of the problem, various models with different depths can be utilized. For example, Kohonen networks, which consist of only one layer, are shallow but useful for biology-related classifications and visualisation~\Citep{suryanarayana2008neural}. In addition to identifying and counting fish and other marine animals, there is enormous potential to apply DL to a wide range of data in coastal ecology~\Citep{grasso2019hunt,marre2020deep}. In the following subsections, we will briefly go through the basics of DNN. 
A glossary of AI terms is summarized in Table~\ref{glossary}.

\begin{longtable}[h]{ p{0.2\textwidth}p{0.7\textwidth} }
\caption{Glossary table.}\label{glossary}\\%
\hline
Glossary \\ 
\hline
\textbf{Accuracy}&Fraction of correct classifications\\

\textbf{Activation}&A non-linear mathematical operation. It is often used to approximate ``turning on'' or ``turning off'' an artificial neuron \\
\textbf{Area Under the Curve}~(AUC)&A summary of the ROC curve that shows capacity of a supervised learning algorithm to distinguish between classes. A perfectly performing algorithm will have an AUC of 1\\
\textbf{Attentions}&A deep learning technique to learn and indicate which sequence in a time series or which regain of image to pay attention to \\
\textbf{Classification}& Categorisation of input data into classes\\
\textbf{Convolution}&Mathematical operation that expresses the amount of overlap of one function as it is shifted over another function\\
\textbf{Convolutional neural network}& A neural network with convolutions, typically used for image classification\\
\textbf{Deep learning / deep neural network}&A neural network with more than one hidden layer\\
\textbf{Encode-decoder}&A neural network that encodes the input data into an internal representation, followed by a neural network that decodes the internal representation, typically to a human readable format\\
\textbf{False negative rate}&The rate of wrongly predicted negative values\\
\textbf{False positive rate}&The rate of wrongly predicted positive values\\
\textbf{Feature extraction }&An operation to elect extract values into feature, typically from unprocessed data\\
\textbf{Features }&Valued characteristics, typically numeric or structural, representing the input data\\
\textbf{Hidden layer }&Any layer of neurons in between the input and the output layers\\
\textbf{Hyper parameters}&User-controlled parameters that influence the model such as number of layers \\
\textbf{Layer}&A set of neurons that takes data as input and typically does a combination of linear~(synapses) and non-linear operations~(activation)\\
\textbf{Loss}&A real number indicating the incorrectness of a single prediction and is typically used to adjust the weights of the neural network \\
\textbf{Machine learning }&Trainable computer programs that learn the representation of data with an aim to predict never-before-seen  data\\
\textbf{Model }& A representation of what a machine learning program has learned. In a neural network, the model is a combined structure consisting of the network and learned weights of the algorithm\\
\textbf{Neural network }& A brain-inspired machine learning technique with an input layer~(features), one or more hidden layers, and an output layer~(predictions)\\
\textbf{Neuron}&A node that combines input data with learned weights and provides a single output \\
\textbf{Object detection}&Recognize the presence of an object instance in a location or area\\
\textbf{Overfitting}&When a model closely predicts the training data but fails to fit testing data\\
\textbf{Weights}&Real values in a neural network in which each parameter individually prioritises each data value, and that are updated in the learning process \\
\textbf{Pattern}&Common trends and regularities in the data such as statistical trends often unique for one category\\
\textbf{Pattern recognition }&Methods to detect patterns in input data \\

\textbf{Precision}&The frequency of true positives among all positive predictions\\
\textbf{Receiver operating characteristic curve~(ROC)}&A graph displaying a supervised algorithm's performance at all classification thresholds. Typically, the relationship between the rate of true predictions and the rate of false predictions\\
\textbf{Recall}&The frequency of correctly identified positive values from all positive values in a data set\\
\textbf{Recurrent network}&Neural networks that connect between nodes to form a directed graph to detect patterns that occur, often over a time series\\
\textbf{Semantic segmentation}&The process of partitioning images into labelled regions\\
\textbf{Supervised learning}&Machine learning that maps an input to a specific, often labelled, output \\
\textbf{Synapses}&Learned weights on the input data for a layer, i.e., how to prioritise the input features\\
\textbf{Labelled training data }&Data used for training the model. It is kept separate from testing and validation data\\
\textbf{Testing data }&Data used for independently evaluating the trained model. It is kept separate from training and validation data\\
\textbf{True predictions}&Model output that corresponds with the correct values\\
\textbf{Underfitting}&When a model has not reliably learned the patterns of the data\\
\textbf{Unsupervised learning}&Machine learning that finds patterns in unlabelled data\\
\textbf{Validation data}&Data used for verifying the model and tuning the hyper parameters during training\\
\hline
\end{longtable}

\subsection{Deep Neural Networks}

All neural networks are function approximators; they mimic the function presented in the training data and adapt to this function through an optimization process. During training, the neural networks' weights, which are many real-valued and connected neurons followed by activations, are updated to match the training data. In more detail, the real-valued difference between the predicted output, $\hat{Y}$, and the expected output, $Y$, is referred to as the loss, which guides the training. For example, $Y$ can be a list of image categories where each value in the vector relates a category to an image, and $\hat{Y}$ is then the neural network's predicted image categories. If the neural network is able to correctly predict image categories, $\hat{Y}$ will be identical to $Y$ and the loss will be zero. The goal of the training process is, generally speaking, to minimize the loss. However, the loss minimization should be done with care since a small loss may indicate that DL has learned specific patterns for each example rather than general trends in the data (i.e., overfitting). To check for overfitting, a separate validation and testing data set is normally employed to independently evaluate the algorithm's performance. 

 A properly trained network has active or inactive neurons that jointly match the training data and minimize the loss. This is analogous to a series of virtual dials that can be turned completely on, completely off, or somewhere in between, indicating the relevance for each feature. During training, the loss for each neuron is propagated backward through the network so that each neuron's contribution matches the product of the weight and a hyper-parameterized learning rate. Hence, each neuron's influence of the loss is matched with a corresponding adjustment of weights, and its adjustment is kept small by the learning rate. When the loss is propagated backwards, the dials are turned slightly in the direction that decreases the loss.

A neural network is considered shallow if it has one layer of input neurons, one layer of hidden neurons, and one output layer. The same network would be considered deep if it had more than one hidden layer, and very deep if it had more than 10 hidden layers. 
Any neuron that is not at the input layer combines a weighted sum from active neurons in the previous layer. The sum is then followed by an activation function for the next layer of neurons.
Despite popular belief, the depth of the DL may not be proportional to the difficulty of the problem that it can solve. It is not always true that deeper networks solve more complicated issues than shallower networks. Some problems can be solved with shallow networks, but in many cases very deep models empirically outperform the shallow ones for image and sound categorisations. For example, a type of neural network called Residual Networks~(sometimes abbreviated to ResNets) often has 18, 34, 50, or 101 layers. Usually, the deeper networks perform better image classification, but occasionally the most shallow network, with 18 layers, is sufficient and even more accurate than the deeper networks~\Citep{aloysius2017review}.




A notable limitation of DL is its dependency on vast amounts of training data. The data requirement typically becomes a significant problem in supervised learning, as a successful application in most cases depends on large quantities of human-classified training examples. This challenge is extensively presented in the marine biology domain, as the limited capacity of trained experts makes extensive and quality-assured labeled training databases hard to acquire. A beneficial property of deep unsupervised learning is its independence of labeled data. However, due to the unsupervised nature, the application area is rather limited in the marine biology domain and has mostly been confined to finding anomalies through re-identification~\Citep{dargan2019survey,ferreira2020deep} and data clustering.

Deep semi-supervised learning has emerged in recent years to mitigate the limitations of supervised and unsupervised learning. Semi-supervised approaches combine training on a small amount of labeled data with a subsequent training phase using large amounts of unlabeled data. In applications where there is often a lack of human-classified training data, semi-supervised learning is especially useful. 

In the paragraphs below, we summarize typical problems relevant to marine ecology where DNN can be utilized as a promising solution.

\subsubsection{Image classification}\label{imageclass}

DNN is the \textit{de facto} standard for machine vision, such as the categorization of images and video files. The most prominent approach among various DNNs is Convolutional Neural Networks~(CNNs), which extract relevant features of an image for subsequent classification by a neural network through a series of two-dimensional mathematical convolutional operations with learnable filters of typical sizes 3$\times$3, 5$\times$5, and 7$\times$7 applied in the image pixels. A CNN trained for classification of images finds the function that best maps the input of pixels to a class, e.g., presence of a fish, plankton or a rope in the photo (Figure \ref{fig:classexample}). Note that the CNN generates small image blocks from the convolutionals of overlapped data within each image. CNN categorizes the image but does not output in which part of the image the object is located. 

The first popularised CNN models were LeNet-1 to LeNet-5 ~\Citep{lecun1995convolutional}, which contain all the basic building blocks still used today. A major advancement came in 2012 with AlexNet~\Citep{NIPS2012_c399862d}, which achieved an error of 15\% of all non-neural network architectures, compared to 26\% previously. These early models suffered from vanishing gradients, meaning that the input data was gradually lost when additional layers were added. This limitation hindered the development of DNN and the performance of the DL models suffered. Later, major innovations included: 1) inception networks~\Citep{szegedy2015going}, which utilized parallel convolutions of different sizes, 2) residual architecture~\Citep{he2016deep}, which added skip connections to allow for an image to both be processed by convolutions and skipped through the network, and 3) Squeeze-and-Excitation networks~\Citep{hu2018squeeze}, which introduced a method to add additional parameters to each convolutional block so that the model could adjust the weight of each block. Each of these innovations has enabled larger, more complex networks.




\subsubsection{Object detection and semantic segmentation}\label{Objectdetection}

Object detection extends CNN models by detecting regions of interest in the image (Figure \ref{fig:classexample}). In addition to classification, a network trained for object detection can output the $x$- and $y$-location, width, and height of the object of interest. This information is then used to draw a boundary box around the object to be classified, e.g. a fish. In this way, a single image can be divided into multiple regions by generating several boundary boxes, allowing for many classes to be classified within a single image. In practice, this means that we can detect and count objects in an image or a video, e.g. the number of fish. 
The approach has been extended even further by pixel-wise detection and classification of the entire image. This approach scales down the image with convolutions and pooling operations, followed by reverse order scaling-up of the same image. This is known as an encoder-decoder architecture~\Citep{girshick2014rich} and allows for categorisation of every region in the image at a high level of detail.

\subsubsection{Individual identification}

A Siamese Neural Network~(SNN)~\Citep{koch2015siamese} is a type of DL model that contains two identical sub-networks with the same layers, hyper parameters, and weights. The neuron weight updates are mirrored and so can be used to find the similarity of the inputs by comparing vector features. An SNN allows us to detect if two images are the same, e.g., two faces are of the same person or two fish photos are of the same fish taken at a different time. Hence, an SNN can classify a new class without re-training the entire network. Other features include robustness to class imbalance (i.e., data is unequally distributed between classes) and learning efficiency in the semantic similarities between images. However, SNNs need more training data and longer computational time than competing networks. 

When training an SNN, a typical loss function used to detect differences in input is a so-called triple loss, in which the baseline input is compared both with a positive and a negative example. A perfectly trained SNN should have a zero loss for the positive example and a loss for the negative example. For example, when detecting individual marine animals, a comparison between pictures of the same animal should have a small loss, while a comparison between pictures of two different individuals of the same species should have a much larger loss. This approach can be used to identify if two pictures include the same individual and verify whether an image consists of an individual that is not part of the training data.

\begin{figure}[h!t]

\includegraphics[width=1\textwidth]{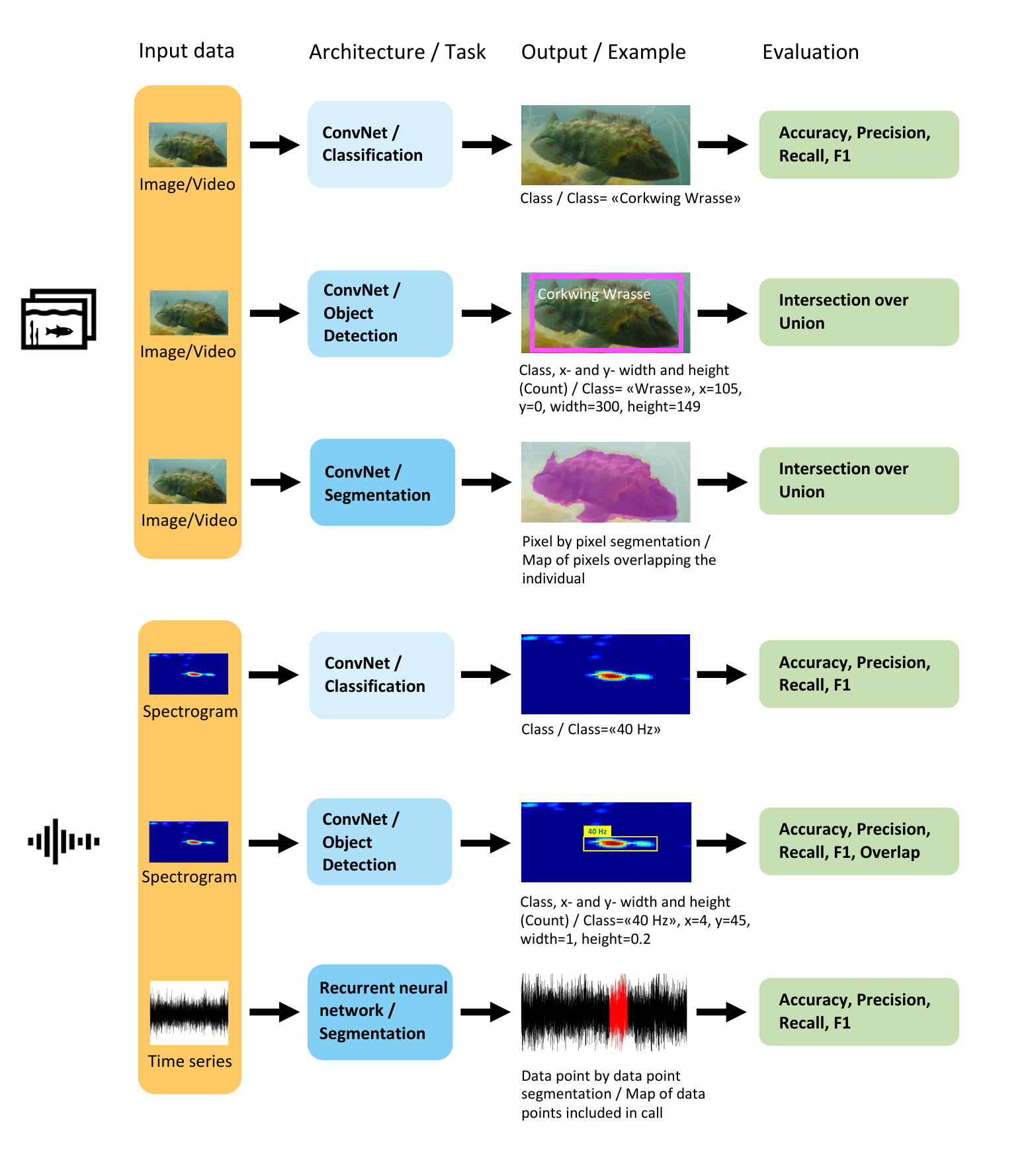}
\caption{Examples of classification, object detection, and pixel-wise segmentation with illustrations of the techniques applied to fish images or audio files.}\label{fig:classexample}
\end{figure}

Figure \ref{fig:classexample} provides examples of classification, object detection, and segmentation and how they are typically evaluated. 

\subsubsection{Audio signal classification}

Audio signal classification~(ASC) is one of the classic and most challenging audio signal processing fields. In brief, ASC comprises capturing appropriate features from an audio sequence and employing these features to distinguish the class that the sequence is most likely to fit. Depending on the application’s domain, one may predict a global signal class with a unique label or a subset of the possible classes with multiple labels. Traditionally, finding appropriate features and designing a suitable classifier are configured as separate procedures. This approach has several drawbacks, e.g. the extracted features might not be optimal for the classification objective and certain features may require prior human knowledge, are difficult to describe precisely, and can be subjective and unstable. As mitigation, DNN-based approaches are developed to perform feature extraction jointly with classification.

To increase the modeling capability, DNN in different structures are usually employed individually or jointly, such as multiple convolutional, feed-forward, and/or recurrent networks $y$~\Citep{goodfellow2016deep}. Feed-forward neural networks have one-way information flow so do not have feedback loops, whereas recurrent neural networks~(RNNs) contain loops. Due to the feedback loops, RNNs can use their reasoning from previous experiences to influence upcoming events. There are different variations of RNN, such as long short-term memory~(LSTM) and gated recurrent units~(GRUs). The most recently developed RNN approaches include attention~\Citep{chaudhari2019attentive, phan2019spatio} and transformer-based strategies~\Citep{tay2020efficient, moritz2020all}. The convolutional concept can be employed together with RNN and attention to improve the performance of the system. For example, an attention-based convolutional RNN model is utilized for environmental sound classification~\Citep{zhang2020attention}. That mechanism adopts frame-level attention to learn discriminatory feature representations for classification. For audio signals that are time series in nature, RNN is especially popular. In some cases, audio is transformed into spectrograms, an image representation of the audio which can be classified using a DNN or CNN. These mechanisms, alone or in combination, can be utilized for audio classification tasks in marine ecology-related applications. Figure \ref{fig:classexample} gives examples of classification, object detection, and data point segmentation with CNN and RNN networks for audio categorisation.

\subsection{Evaluation criterion }

\begin{figure}[h!t]

\includegraphics[width=0.99\textwidth]{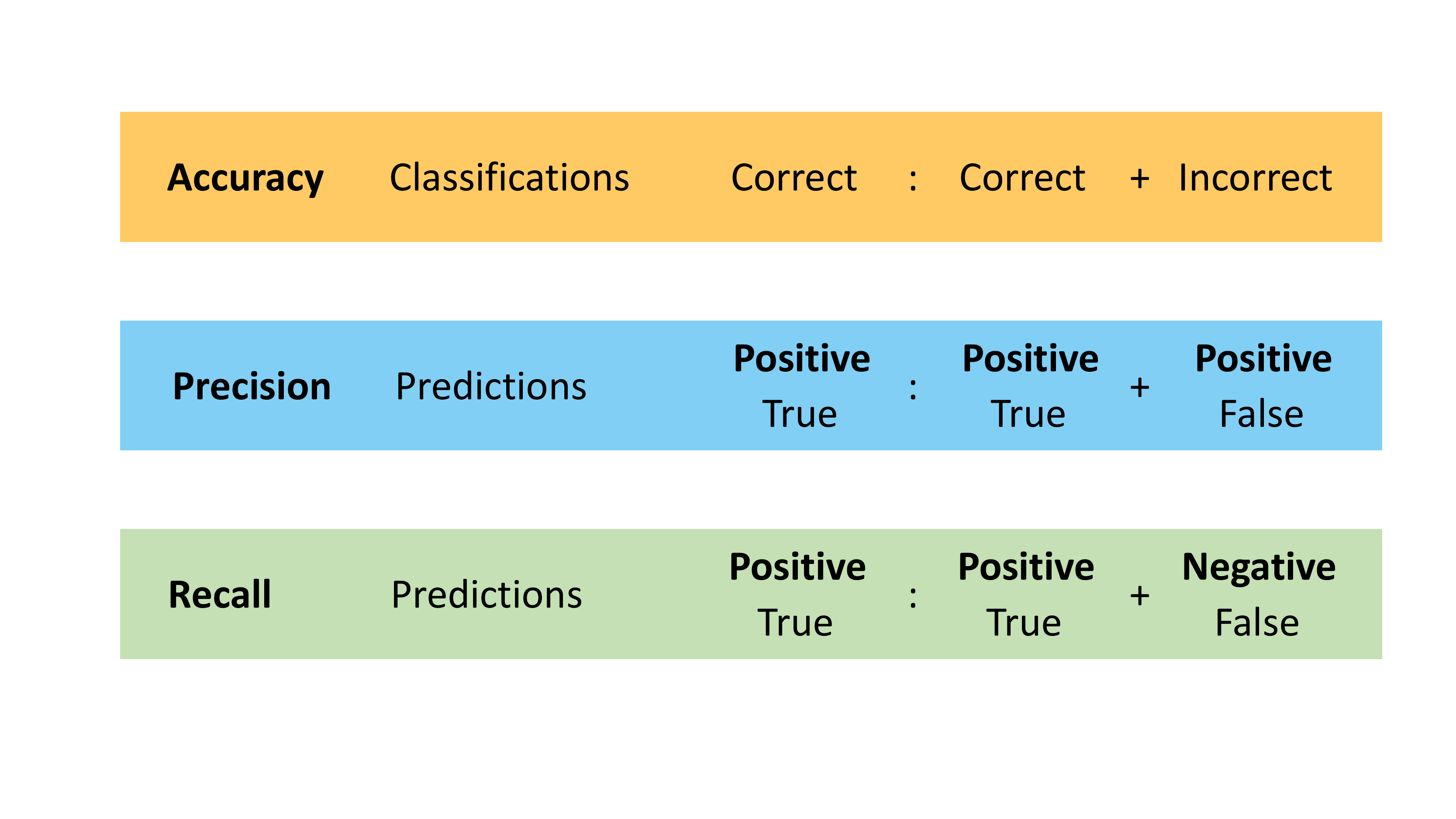}
\caption{Evaluation metrics, accuracy, precision, and recall, for classifications and predictions.}\label{fig:evaluation}
\end{figure}

To evaluate the performance of a trained model, different parameters are utilized by the different approaches, such as accuracy, precision, and recall (Figure \ref{fig:evaluation}). 
Accuracy is the ratio of correct classifications to the total number of classifications. Precision for the positive predictions is the ratio of true positive predictions over the sum of true positive and false positive predictions. The same concept applies to the precision of negative prediction. Recall is the ratio of true positive predictions over the sum of true positive and false negative predictions. A result of a DL algorithm may be precise but not accurate when results are biased but with small variance. A DL algorithm is considered valid if it is both accurate and precise. 

For example, if the expected output $Y$ is 5 images of cod and 5 images of trout, and the predicted output $\hat{Y}$ correctly identifies all cod and only 4 of the trout, with one trout wrongly identified as cod, the algorithm is correct 9 out of 10 times, yielding an overall accuracy of 90\%. In this example, the precision for cod is 100\%, i.e. all cod were predicted as cod, but only $\frac{5}{6}=83\%$ for trout, i.e. for all the predicted trout, only 83\% are actually trout. The recall for cod will be 100\%, i.e. the algorithm identifies all cod, but the recall for trout will be $\frac{4}{5}=80\%$, i.e. the algorithm only identifies 80\% of the trout.  

The parameter used for performance evaluation depends on the data. Accuracy is most suitable if the data set is balanced, meaning an approximately equal number of examples in each class, and where false positives and false negatives have similar implications. But if the data set is imbalanced, which is typical for ecological data~(e.g., some species are more common than others), precision or recall are better. A high precision relates to a low false-positive rate, whereas a high recall relates to how well the model detects the class in the total data set. The F1 score, a unified metric, is a weighted average of precision and recall and therefore encompasses both the false positives and false negatives. A rule of thumb is: if in doubt, evaluate your algorithm with the F1 score. 


\subsection{Data}
There is no universally right answer as to how much data is needed --- generally, the more data, the better. Learning an intricate pattern requires more data than learning a simpler one. For example, for a DL to classify an image as either a sea trout or another fish species with clear morphological differences, such as a cod, it may achieve a near-perfect separator with relatively few samples. However, more data are likely to be required for a model to learn to distinguish sea trout from a closely related species with similarities in appearance, such as salmon, simply because that is a more complex task to learn.

Mitigation for the lack of data means using an existing model with weights pre-trained using other data sources, such as the ImageNet database~\Citep{imagenet_cvpr09}. The typical approach is to first train with an available, sizeable dataset and subsequently train with a smaller but more relevant dataset. In this way, the learning algorithms find the general image patterns from a big dataset~(e.g., shapes, species patterns, face patterns) and the individual differences from the smaller dataset.  

For a classification or object detection task, the dataset needs to be labeled~(sometimes referred to as annotated), usually by a human expert (e.g., an ecologist). The labeled data is often referred to as the $Y$ vector. An accurate classifier algorithm should correctly map the input, known as the $X$ vectors~(e.g., images) to the appropriate $Y$ vector~(the labels). These predicted labels are often referred to as the $\hat{Y}$ vector, regardless of whether the predictions are correct~($\hat{Y}$ matches $Y$), or incorrect~($\hat{Y}$ does not match $Y$).

The labels for a classification task are distinct for each input variable, such as a species of fish for each image. This requires manual categorization and labelling of a large set of images. For object detection and semantic segmentation, the labels must also indicate where in the image the object of interest is located. In the case of audio input for RNN and CNN classification, the start and stop times of all events of interest must be labelled in order to segment the data into relevant categories. If object detection is used on spectrograms of audio, the frequency bands must also be labelled, encasing the contours of interest in the spectrogram. As is the case with images, existing labeled datasets also exist for audio, which can be used for training when data is otherwise limited (e.g., the DCLDE 2015 data set for baleen whale social calls~\Citep{huang2016automated}).   

A labeled dataset is divided into three separate datasets, as illustrated in Figure \ref{fig:AI}: training, validation, and testing. The training set is used to train the model, meaning that it tries to find an approach to map the training set's input vector $X$ with the training set's correct labels $Y$. The validation set follows and is first used to check whether the algorithm can map the validation set's input vector $X$  with the validation set's correct labels $Y$, which is separate from the training set vectors. Finally, the test set is then employed to blindly verify that the data set with its own input vector $X$ maps to the labeled data set $Y$. This test is the final check of how well the algorithm can classify. 





%

\section{Established cases: identification and quantification of marine biodiversity}\label{sec:Established}

\begin{figure}[h!t]

\includegraphics[width=0.99\textwidth]{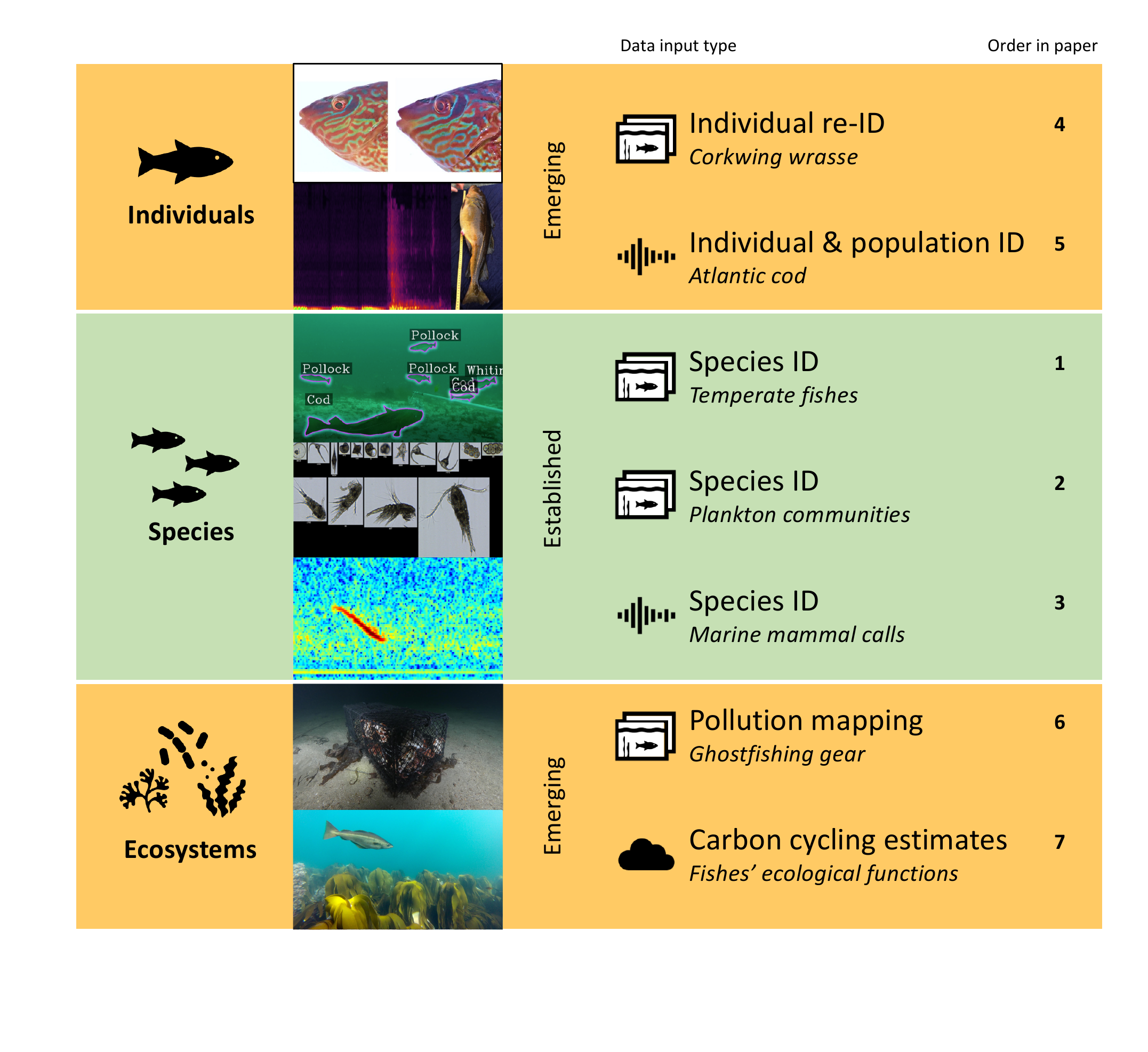}
\caption{Established and emerging cases for deep learning in marine biology, from individuals to ecosystems. Data input type icons represent images/video (cases 1, 2, 4, and 6), audio (cases 3 and 5), and large-scale environmental monitoring data that is often stored on remote servers (i.e., "the cloud"; case 7). Kelp forest photo (bottom) credit: Frithjof Moy/Havforskningsinstituttet.}\label{fig:OverView}
\end{figure}


The application of DNN provides an alternative to laborious or repetitive manual tasks, such as processing data from underwater recording equipment. The following section presents three cases in ecological research where deep learning is already used to alleviate data processing and is likely to become the method of choice. These cases exemplify the DL methods described in Section 2:  
\begin{itemize}
\item Image and video classification to identify fish species and track movement, 
\item Image-based analysis for monitoring of plankton, and
\item Passive acoustic monitoring of whales.
\end{itemize}

\subsection{Case 1: detection, classification, and tracking of fishes in images and videos}\label{ObjectdetectionFish}

Monitoring of fish populations and communities is a central activity within marine management and conservation. Traditional sampling methods to track population trends, estimate abundance, and to infer movement patterns of fish have relied on studies that involve animal handling~(i.e., fishing gears, individual tags, biologgers). These methods are not only invasive, but also time consuming. Developing and applying passive ways to both obtain the necessary data and to speed up analysis are therefore imperative. Today, automated detection, classification, and tracking of small-scale movements of fish through images and video are made possible with deep learning, an application well suited to this task.

When selecting AI approaches for monitoring, consider that a real-life underwater scenario typically involves multiple fishes present in the same image, which precludes the use of standard classification techniques. A solution to this problem is to introduce object detection before classification. The object detection step discriminates between individuals within an image and separates them, and in this way, prepares the image data for classification. Object detection and classification can be two completely separate steps in a pipeline~\Citep{knausgaard2021temperate,connolly2021improved}, or integrated as part of an object detector, such as YOLOv1-YOLOv4~\Citep{redmon2016you,bochkovskiy2020yolov4,yang2021deep,shin2021deep,jalal2020fish}. 

Detecting and counting species from still images and videos is relatively straightforward using standard DL object detection algorithms, as described in Section \ref{AIoverview}. However, a challenge with setting up a detection algorithm is that well-established object detection training datasets, such as Coco~\Citep{lin2014microsoft} and ImageNet~\Citep{deng2009imagenet}, include few images within the category of each species of fish and with very little variation in the background. Thus, the applicability of such datasets becomes somewhat limited. To increase the precision of detection suited to the specific use, one should instead train the DNN with images of fishes in their natural environment. Collecting and labelling relevant image and video data is therefore central to building a high-performance and robust fish detector. Public datasets are currently an integral part of this research, particularly for fish detection and species identification (e.g., Fish4Knowlege~\Citep{fisher2016fish4knowledge}, datasets of temperate fish species ~\Citep{knausgaard2021temperate}, and across species, location, and depths, as in NOAA fishery datasets~\Citep{link2015noaa} and the OzFish dataset~\Citep{ditria2021annotated}). The best performance by AI in species identification (i.e., classification) is achieved with a specialized CNN that only classifies species without detecting at the same time. The squeeze-and-excitation-based CNN presented in~\Citep{knausgaard2021temperate} reached classification accuracy of 99.27\% on the  Fish4Knowledge dataset~\Citep{fisher2016fish4knowledge} and 87.74\% on a second temperate species dataset.

Marine researchers often collect videos rather than still images and are interested in tracking the same animal across consecutive frames to obtain information on behaviour~(e.g., to estimate swimming speed,~\Citep{Beyan2015}), or to ensure that the same fish is not counted multiple times~\Citep{lopez2021slow}. To continuously follow a moving object's position in a video sequence, such as a swimming fish, object tracking can be used. One way of implementing tracking is to use a detection algorithm that feeds another tracking algorithm with position data. When tracking multiple objects (e.g., a school of fish), a track association decision needs to be made for each object (e.g., each individual fish). Thus, a complete tracking system typically consists of a detection algorithm, association of detection with tracks, and the actual tracking algorithm. In practice, tracking commonly involves Kalman filters or other recursive estimators to enable efficient dynamic tracking of objects ~\Citep{ristic2004beyond}, including specific fish~\Citep{2021_BarreirosZebraFishTrackingKalman}. Another emerging approach is to let DL solve the entire multi-class tracking problem in one step~\Citep{ciaparrone2020deep}. This one-step approach typically results in a more homogeneous system, but with less fine-scale control than when applying well-understood recursive estimators. Further, a fully integrated CNN-tracking approach leaves less room for the user to include \textit{a priori} information on expected fish dynamics and behaviour. A CNN-only approach will, however, completely avoid the meticulous tuning requirement of mathematical models and Kalman filter parameters.


We see DL as an essential building block for automating image and video analysis where the goal is to quantify, classify, and track fish. DL can either be used in a modular pipeline with separate steps for detection~\Citep{knausgaard2020temperate}, association, and track building, or as a complete solution to a multi-object tracking problem~\Citep{yang2021deep,shin2021deep,jalal2020fish}. As these DL tools are adaptable for use with different ecosystems or species by virtue of the training datasets used, the potential for AI in monitoring is great.

\subsection{Case 2: image-based analysis for plankton monitoring} \label{plankton}
Plankton is a highly diverse group with very different morphologies and sizes ranging from submicrons to a few centimeters, or even a few meters~\Citep{lombard2019globally}. Plankton are responsible for about 50\% of global primary production~\Citep{field1998primary}  and constitute the base of many marine food webs. Some species serve as bioindicators of ecosystem health, while others can form toxic blooms with adverse impacts on other marine life, including commercially important fishes. Therefore, tracking seasonal, interannual, and spatial changes in plankton composition and abundance is central to coastal monitoring. As such, an ever-increasing volume of plankton images is generated for monitoring each year. Various AI approaches have been developed to analyse this data and reduce manual processing. Plankton identification and counting are arguably some of the most useful examples of DL in marine biology. The ultimate goal is fully automated plankton classification without human biases~\Citep{culverhouse_natural_2007}. This bias is not trivial, as human experts can only achieve 67-83 \% self-consistency during a difficult classification task~\Citep{culverhouse2003experts}, although accuracy is much higher ($>$90\%) when working with natural plankton samples with many taxa which have variable classification difficulty~\Citep{luo2018automated}. 

Several systems for image acquisition and AI analysis of plankton are commercially available ~\Citep{lombard2019globally}, including \textit{in situ} ~(e.g., Imaging FlowCytobot, VPR, IISIS) and those that image samples, fixed or fresh, on research vessels or in the laboratory~(e.g., ZooCam, FlowCam). All approaches share the same basic principles: pictures are taken of the sampling volume and the objects are segmented~(i.e., into individual organisms). Each segment is then classified into one of several pre-defined classes, typically taxonomic or functional groups, but living organisms are always separated from non-living particles. Besides the predicted classification, the algorithms can extract object features~(e.g., length, width, equivalent spherical diameter), and therefore information on plankton community structure and function~(e.g., normalized biomass size spectra~\Citep{wang_vision-based_2020}). Seasonal and interannual variability in plankton abundance and composition obtained using these image-based DL methods is comparable with traditional microscopy~(e.g., FlowCam,~\Citep{alvarez_routine_2014}).

Initial plankton classification models were based on statistical approaches but soon transitioned into machine learning solutions~\Citep{kerr2020collaborative, luo2018automated}, including algorithms that classified plankton based on object features such as size or edge, for example Support-Vector Machine and Random Forest algorithms ~\Citep{fischer_return_2020, faillettaz2016imperfect}. These algorithms reach 70-90\% accuracy in classification for the most abundant plankton groups, but rare or cryptic species can still be a problem. Simpler classifiers cannot extract the object features from the raw data and instead require these to be manually defined by ecologists, a cumbersome process. CNNs to overcome these issues are being proposed, such as collaborative CNNs with configurations to deal with class imbalance~(e.g., where one type of plankton is much more frequent than another) ~\Citep{kerr2020collaborative} or when the environment dynamically changes (dataset shift) using a supervised quantification scheme ~\Citep{orenstein_semi-_2020}. These CNNs achieve state-of-the-art ~90\% classification accuracy when classifying independent test sets~(e.g., 97\% accuracy classifying 0.1 million FlowCam images ~\Citep{kerr2020collaborative}), although accuracy decreases with very many diverse images~(e.g., 83\% accuracy for 52 million zooplankton images from IISIS ~\Citep{briseno-avena_three-dimensional_2020}).  
Other approaches to improve accuracy of conventional CNNs are through inclusion of context data~(e.g., sampling location and time) in the classifier~\Citep{ellen_improving_2019}, using unsupervised clustering of data~\Citep{schroeder_morphocluster_2020}, or combining CNNs with Support-Vector Machine (SVM) classifiers~\Citep{cheng_method_2020}. 

DL enables a whole new approach to plankton coastal monitoring by (semi-) automatic analysis of samples either \textit{in situ} or in the lab~\Citep{wang2019advancing}. DL is used to monitor long-term, seasonal, and spatial changes in taxonomical groups~\Citep{briseno-avena_three-dimensional_2020} and size spectra ~\Citep{wang_vision-based_2020, yu_automated_2016}, to track plankton that serve as bioindicators of ecosystem health~\Citep{uusitalo_semi-automated_2016}, or as an early-warning system for harmful algal blooms that impact higher trophic levels and, ultimately, humans~\Citep{gorocs_deep_2018,orenstein_scripps_2020}.  However, DL cannot replace a taxonomist for difficult identification tasks~(e.g., identification of certain species or life stages of zooplankton or larval fish), and as such are not yet adequate for studies that require high taxonomic resolution. Experts are also required to create training sets and validate the results. However, manual hours can be reduced if training sets and analysis pipelines are made publicly available~\Citep{li_developing_2020, 875nf10421, schmid_moritz_s_2021_4641158}, as well as through the creation of global databases and training sets~(e.g., Ecotaxa ~\Citep{picheral_ecotaxatool_2017}). Ultimately, the combination of traditional physical plankton sampling with autonomous platforms that combine image-based data with data from other sources~(e.g., genomics, acoustics, pigments) appears to be the best way forward for coastal plankton monitoring studies~\Citep{gorsky_expanding_2019, lombard2019globally}.

\subsection{Case 3: passive acoustic monitoring of whales}\label{whale}
The use of long-term underwater passive acoustic monitoring~(PAM) recording has grown in the last couple of decades to become an indispensable tool for investigating relative population trends and temporal and spatial migration patterns of a wide range of whale species~\Citep{wiggins2016long}.

For many years, the standard procedure for detecting and classifying whale calls from PAM recording has been to retrieve the sound recording, use a software package like Triton~\Citep{ wiggins2007high} to create spectrograms lasting 1-2 minutes, then have the spectrograms manually scanned for call contours by a trained data analyst. This method is not only highly labor-intensive, as PAM recording can cover months, if not years, but the results are also subjective~\Citep{ baumgartner2011generalized}. As many whale calls are highly stereotypical, algorithms like matched filtering~\Citep{giannakis1990signal} and spectrogram correlation ~\Citep{ mellinger1997methods} have successfully been developed for automated call detection. However, these methods tend to work poorly on calls with more variability in frequency modulation. Hence, manually scanning spectrograms continues to be used for many call types. 

The manual procedure of visually scanning spectrograms for known call contours is very similar to the image classification process explained in Section \ref{imageclass}. Further, sound classification using deep learning is becoming well established outside of marine aquatics~\Citep{piczak2015environmental,mushtaq2021spectral,Sharma2020}, which has led to significant interest in using CNN for automated whale call detection.

Among the whale calls recently being investigated using CNN are those of the beluga whale~(\emph{Delphinapterus leucas}) with an AUC of 0.9906~\Citep{ zhong2020beluga}, North Atlantic right whale~(\emph{Eubalaena glacialis}) with an AUC of 0.902 ~\Citep{ shiu2020deep}, killer whales~(\emph{Orcinus orca}) with an AUC of 0.9523  ~\Citep{ bergler2019orca}, and sperm whales~(\emph{Physeter macrocephalus}) with 99.5 percent accuracy in detecting sperm whale clicks in 650 spectrograms ~\Citep{ bermant2019deep}. A drawback of CNN classification without object detection is that it does not relay information about where in the image an object is located. For example, when examining spectrograms where the $x$ axis is the timeline, no information is included about the call's specific time, nor the number of calls, thus the CNN serves as a ``presence'' identification tool only. A work-around for this issue has been to make the spectrograms very small, covering only a short timeline (e.g., two seconds)~\Citep{ bergler2019orca}. When creating a spectrogram, there needs to be an overlap between two consecutive spectrograms. Otherwise, a call located at the intersection of two spectrograms might be missed. Using short spectrograms combined with these overlaps can increase the redundant data up to 20\%~\Citep{bergler2019orca} and thereby increase the computational cost at a similar level. 

Object detection, as described in Section \ref{Objectdetection}, would solve these issues for whale call detection. For example, a custom-made region-based CNN for detecting regions of interest in combination with a transformed pre-trained CNN for further classifying the regions of interest was successfully trained and tested on the highly variable D call emitted by blue whales and 40 Hz calls emitted by fin whales~(\emph{Balaenoptera physalus})~\Citep{doi:10.1121/10.0005047}.

Looking to the future, use of AI generally, and DL specifically, in automated detection of whale calls in PAM recordings will undoubtedly benefit from the recent developments in networks architecture search (NAS) algorithms ~\Citep{sun2019completely}. This new technique of automatically developing network architecture from prefabricated blocks will cut down significantly on the work needed to adapt networks to fit specific species and calls, and make CNN more accessible for whale researchers. A general move from using CNNs to perform image recognition on spectrograms extracted from the PAM to using DL directly on the PAM is also anticipated. This can be done via recurrent networks like long short time memory networks ~\Citep{hochreiter1997long} or a recently developed type of network called the transformer~\Citep{vaswani2017attention}.

\section{Emerging cases}\label{sec:roadahead}

A common theme of the established cases mentioned above is that they replace tasks currently conducted by humans - where using DL can reduce costs, labour, and sometimes improved accuracy compared to human analysts. However, DL has the capacity to be applied to solve more complex tasks, detecting patterns in visual and acoustic data that are difficult for humans to reliably detect or discriminate. In this section, we illustrate novel research avenues in which we predict DL will be successfully applied in the near future.

\subsection{Identifying and characterizing individual phenotypes}

\subsubsection{Case 4: visual re-identification of individuals in wild fish populations}\label{REID}

Methods for individual identification are needed to answer many questions in animal behavior and ecology, such as growth, movement, and survival inferred from capture-recapture studies~\Citep{Clutton-Brock2010}. Currently, the most common approach is to mark animals with various physical identifiers to recognise individuals upon re-sight or re-capture, such as leg rings on birds, number scratching or paint on reptiles, or lip tattoos on larger carnivores. In marine and freshwater systems, capture-recapture studies on fish are most often performed using external number tags or radio-frequency identification~(RFID) tags~\Citep{Pine2003}. However, trapping and tagging surveys are often costly, logistically challenging to conduct, and are intrusive to the animals.

A less invasive and more practical way forward for data collection is to use images or videos from wildlife cameras and perform DL image analysis by taking advantage of natural markings that make individuals identifiable~\Citep{Schneider_2019}. Like humans, many animals have unique features about their individual appearance, such as intricate patterns of spots and stripes on the skin, fur, or feathers. A trained computer vision algorithm can distinguish between individuals as different classes, even when the identifying features are highly complex. CNN networks have been trained to recognise individuals~(individual re-identification [Re-ID]) from photos of animals across many taxa, including birds~(e.g., 93.6\% accuracy ~\Citep{Ferreira_2020}), turtles~(e.g., 95\% accuracy ~\Citep{Carter_2014}), and terrestrial and marine mammals~(e.g., 92.5\% accuracy ~\Citep{Schofield_2019}). Many fish species also have solid visual pigmentation; stripes, spots, or mosaic in contrasting colors that can be clearly seen in images and video surveys ~\Citep{Dala_corte_2016, Hau_2019, Mucientes_2019}, particularly coastal fish like the corkwing wrasse~(\emph{Symphodus melops}; Figure \ref{fig:OverView}). Therefore, development of Re-ID has potential to replace physical tagging for individual identification of teleost fish, and would also be of great value for monitoring, as it could be used to assess individual movement, behaviour, and growth. Re-ID could also solve the problem of double counting when individuals re-enter the field of view, thus improving video-based monitoring of abundance ~\Citep{Aguzzi_2015, Campos_2018, Perry_2018}

As far as we are aware, Re-ID by CNN has not been tested in wild populations. One of the challenges preventing the widespread development of AI-based Re-ID is the need for photos or videos of known individuals, independently validated with high certainty, for the training and validation of the algorithm. One solution to this problem is collecting data by using remote detection systems, such as RFID technology, to identify individuals tagged with passive integrated transponders~(PITs). By combining PIT-tagging with RFID and synchronized underwater cameras, a large, automatically labeled dataset of many individuals could be created over a relatively short time~\Citep{Ferreira_2020, Schneider_2019}.

\subsubsection{Case 5: inter- and intra-individual variability in fish vocal communications}\label{Vocal}

    
Acoustic communication is a fundamental component of animal life, especially for aquatic species for which visual cues are not as effective~\Citep{tessler2017deep}. For example, many fishes hear their species mating choruses from several kilometers away~\Citep{winn1964biological}. Subtle variation in complex acoustic signals is challenging for humans to detect or interpret. Furthermore, using algorithms to detect patterns that defy human perception has technological limitations, including processing high volumes of noisy, real-time acoustic data. Using algorithms to detect acoustic signaling presents the additional challenge of source identification in moving animals. However, advances both in audio recording technologies and in DL algorithms that can detect and classify acoustic signals in natural settings have opened up new systems for study, both on land and at sea~\Citep{parsons2009localization}. These technologies unlock the potential for understanding inter-and intra- individual variation in acoustic communication of fishes. 
    
    Marine mammals are relatively well studied in this respect, as vocalizations can be classified at the species, population, and even individual levels~(e.g., Case 3). However, understanding of the diversity of fish vocalizations and how these vary within species is poorly understood. Moving beyond species-level to population- and individual-level classification of vocalizations is necessary to understand the ecological and evolutionary consequences of acoustic communication in fishes and the potential impacts of anthropogenic noise pollution on them. Likewise, for better understanding of intra-individual variation in communication, which is necessary for understanding the role of vocalization in fish behavior and personality.
    
    A prime example is Atlantic cod~(\textit{Gadus morhua}), which use drumming vocalizations during social interactions, particularly during mating~[Brawn 1961]. Yet, our understanding of inter-and intra-individual variation in drumming is limited. There is potential to catalog individual variation in sound production using DL algorithms~\Citep{deng2014deep}. Fine-scale individual variation in fish sounds, especially without \emph{a priori} knowledge, is beyond human perception. Thus, automating this task requires DL approaches that do not rely on labelled training sets. Specifically, CNNs can detect and classify fish sounds by implementing a transformer network~\Citep{deng2014deep}. Transformer networks work solely on optimized attention and are currently state-of-the-art in translation tasks.  The transformer network is rapidly replacing recurrent neural networks~(RNN) previously used for this kind of task, as it solves two of the problems inherent in RNNs: 1) long computing times due to serial processing and 2) vanishing gradients (see Section 2.1.1).

\subsection{Ecosystem}

\subsubsection{Case 6: ghost fishing gear detection}\label{ghost}

   When fishing gear is lost, the continued mortality of fish, crustaceans, and other species caught in the gear is termed ghost fishing~\Citep{brown2007ghost}. The problem is widespread and high rates of fish trap loss are reported~\Citep{vadziutsina2020review}. Using DL to detect and locate lost gear can greatly increase the efficiency of clean-up efforts, as human effort could then focus on retrieving gear (e.g., using remotely operated vehicles). Detection of ghost fishing gear has been achieved using side-scan sonar for data acquisition followed by feature cloud generation, which involves looking for objects in an image by identifying areas of high entropy, then clustering and noise reduction to separate the objects from noise by looking for clusters of the identified areas ~\Citep{labbe2020unsupervised}. 
   
   The next step is using autonomous object detection to extract the location of lost fishing gear. The detection of lost fishing nets using a towed underwater camera followed by automatic object detection has been achieved with a region-based CNN~(R-CNN)~\Citep{POLITIKOS2021111974}. In that study, fishing nets were detected with higher precision than any other type of marine litter. Detection of more types and features of fishing gear is of interest to researchers and clean-up efforts (e.g., whether the feature detected is a trap, fyke net, or ropes). Image classification may be an effective approach to provide this level of detail, where low resolution images are not usually a hindrance for successful image classification. As well as video, side-scan sonar on autonomously operated vehicles could provide the data needed for this approach. Towed underwater cameras may represent a low-cost option for data collection, whereas autonomously operated vehicles equipped with side-scan sonar represent a high-cost option.

\subsubsection{Case 7: carbon cycling by fish}\label{carbon}

    The ocean sinks approximately one third of greenhouse gas emissions out of the atmosphere, including carbon dioxide. The ocean carbon sink is driven by a physical and a biological pump. As well as plankton and bacteria, fishes contribute to the biological pump, with recent estimates suggesting 16 percent of sinking carbon could be due to fishes~\Citep{saba2021toward}. However, the role of fish in the biological pump is not well understood~\Citep{martin2021integral}. The data on fishes required to improve our understanding relates to metabolic use and excretion of consumed carbon and other nutrients; properties of carbon and nutrient outputs and their fate in the environment; habitat use and connectivity of ecosystems; and physical interactions with extrinsic carbon and nutrients in the environment. As well as advancing knowledge of the role of fishes, this knowledge could inform effective management approaches to maintaining or restoring ecosystem carbon function. As an emerging field, zoogeochemistry has the advantage that much of the relevant data are already published for other purposes. For example, metabolic rates and behavioral data is already published for many commercially important species through fisheries and climate change research. Using AI in this field has the potential to expedite a better understanding of fishes ecological functions, effects of human disturbance, and therefore potential management of important carbon sink habitats. Here we present a few of the options available to apply DL to zoogeochemistry research. 
    
    In habitats where visual sampling is possible, video images could be used with object detection, classification, and tracking to identify the presence or absence, behavior, and features of particles from fishes and their short-term fate ~(e.g., defecation, spawning, and whether material reaches and settles on the sea floor). This could inform estimates of the volume of carbon transferred into or out of a habitat by fishes, and the short-term fate of the carbon or nutrient they release. Methods that use AI computer vision to determine the connectivity of fish populations can also be of value in estimating carbon flow ~\Citep{lopez2021slow}. The long term fate of carbon and nutrients depends on physical, chemical, and biological conditions of the environment. Graph networks have recently been used to simulate the physical behavior of materials~\Citep{sanchez2020learning}. This technology has potential application to estimating the probable fate of carbon and nutrient outputs through simulations that combine oceanographic data with features of the carbon released by fishes. With many variables to consider, recent approaches to assessing carbon contained in sediments in different habitats include a combination of survey~(acoustic and image-based) and bathymetry data, modeling, and remote ground-truthing~\Citep{hunt2020quantifying,wilson2018synthetic}. The current approach is manual, but there is potential for AI application to link habitats to carbon fates and make spatial and temporal estimates on cycling and sinking of carbon and nutrients. Graph networks~\Citep{sanchez2020learning} could be applied to generate probable long-term fates of carbon and nutrient outputs using simulations based on video observations and environmental parameters such as season, temperature, currents, and maps of habitat type.

    As has been mentioned in earlier cases, biological data for fishes is partially or fully available for commercially targeted species in online databases (e.g. Fishbase). Such databases have been used to generate estimates of nutrient output from fishes, such as nitrogen and phosphorous~\Citep{schiettekatte2020nutrient}. AI can be trained on these databases to estimate ecological and behavioural carbon flows, including on food webs and habitat use. This training could then be applied to generate estimates for species where ecological data is limited, such as deep-sea fishes. The research needs for deep sea fishes are urgent as commercial interest is increasing at the same time as the significance of these species in moving carbon from surface waters to the deep sea is beginning to be explored, but data collection methods are expensive, time consuming, and patchy~\Citep{martin2020oceans, bohan2011automated, lyubchich2019using}. In this instance, DL could be used to detect probable carbon flows by using logic-based machine learning.

\section{Discussion}\label{sec:Discussion}

We are entering a new era in ocean research and management thanks to new technological developments in observational methods combined with AI-supported data analysis. Data collection, processing, and interpretation are at the core of ecological studies and biodiversity monitoring. Scientists are increasingly relying on indirect observations from various sensors generating large and complex data sets, especially in the aquatic environment. Thus, we envision that within a decade, marine researchers will firmly integrate AI and ML in data collection and analysis within most sub-fields of applied marine biology. This development will only continue to accelerate with new generations of biologists better educated in computer science and informatics~\Citep{weinstein2018}.

Non-human, autonomous, and remote platforms such as cabled observatories, autonomous underwater vehicles or gliders, and ships of opportunity will have a pivotal role in ocean monitoring~\Citep{10.3389/fmars.2020.00697}. These platforms will record continuous, real-time information on water physics, chemistry, community composition, and biomass of plankton, fish, and other marine species. For example, long-term monitoring of harmful bloom-forming plankton species can be achieved using inexpensive image technology anchored to piers~\Citep{gorocs_deep_2018, orenstein_scripps_2020}. Similarly, changes in whale population trends and migrations can be investigated using passive acoustic monitoring~\Citep{szesciorka2020timing}. These methods are likely to decrease reliance on manual analysis or direct sampling via more invasive, expensive, time-consuming, or labour-intensive traditional approaches. This new way of observing the ocean will generate large volumes of data that will only be feasible to analyze with the help of AI. Therefore, AI will play a key role in making routine processes more time-efficient and alleviate the manual work required. As an example, a trained data analyst currently needs 50 to 350 workdays to manually scan one year's worth of PAM recordings for whale calls. In contrast, the same task can be accomplished by a trained neural network in approximately four days~\Citep{bergler2019orca}. Fully automated coastal monitoring systems will be faster and more efficient at detecting changes of interest, such as necessitating warnings to the public where toxic algae are abundant and enabling redirection of boat traffic where whales are moving across shipping routes. Altogether, this monitoring information will be valuable in the development of indicators and in integrated assessments to support ecosystem-based management~\Citep{10.1093/icesjms/fsw230}. It is important to emphasize that expert work will always be needed to create and correctly label training sets and revise the automated analysis, such as when new species enter the system. However, this anticipated demand emphasizes the need to develop interdisciplinary skills in researchers at all career stages, as well as the skills required to form fruitful collaborations~\Citep{https://doi.org/10.1111/gcb.14168}. 

Collaborative work based on open access and sharing culture~(from model configurations to training sets) will be essential to advance this future. While this is a common practice within AI communities, the culture of marine science is not as open. However, funding agencies, publishers, and institutions are increasingly enforcing open access for data generated via public funds. The FAIR Principles for scientific data management and stewardship are now widely adopted~\Citep{wilkinson2019fair}. These emphasize improving the access, utility, and reuse of data by machines in addition to individual researchers. As such, they may play a vital role in applying AI to the marine domain. Some collaborative initiatives are underway to create global databases for plankton and benthic images and training sets~(e.g., EcoTaxa~\Citep{picheral_ecotaxatool_2017} and BIIGLE~\Citep{10.3389/fmars.2017.00083}), as well as pipelines~\Citep{875nf10421}. Ultimately, we envision libraries of images, videos, metadata and more available globally, similarly to the open access GenBank database for sequence information and associated metadata for genetic material hosted by the National Center for Biotechnology Information (NCBI) in the United States.

\section{Conclusions and future directions}\label{conclusions}

We have provided examples of how image and audio analysis are already used to analyze marine biodiversity distribution and dynamics in non-invasive ways, emerging applications of AI, and a look at what the future of AI in marine ecology requires. The United Nations Decade of the Ocean has just started, with the aim of achieving “a healthy, safe, and resilient ocean for sustainable development by 2030 and beyond”. We have shown that AI will be key to achieve this goal by developing new technology to uncover new aspects of and potential threats to marine ecosystems' structures and functions, thereby informing EBM. This new knowledge will directly address several of the key challenges identified for the Decade, from effective EBM and biodiversity conservation, to creating a digital representation of the ocean and delivering data, knowledge, and technology to all. The Decade of the Ocean initiative promotes global cooperation and interdisciplinary efforts at all levels, which are at the core of how AI-linked marine studies will progress. Where researchers have the opportunity to gather large amounts of complex ecological data, unfamiliarity with AI jargon and the latest developments should not prevent collaborations with data and computer scientists to support EBM of ocean resources during this time of rapid change.

\section*{Acknowledgements}
Morten Goodwin is supported by the Norwegian Research Council HAVBRUK2  innovation project CreateView Project nr. 309784. Rebekah A. Oomen is supported by the James S. McDonnell Foundation 21st Century Postdoctoral Fellowship. Susanna Huneide Thorbjørnsen is supported by Handelens Miljøfond.

\bibliographystyle{unsrtnat}
\bibliography{references}

\end{document}